\documentclass{article}

\PassOptionsToPackage{numbers, compress}{natbib}

\usepackage[preprint]{neurips_2026}

\usepackage[utf8]{inputenc}
\usepackage[T1]{fontenc}
\usepackage{hyperref}
\usepackage{url}
\usepackage{booktabs}
\usepackage{amsfonts}
\usepackage{amsmath}
\usepackage{wrapfig}
\usepackage{nicefrac}
\usepackage{microtype}
\usepackage{xcolor}
\usepackage[pdftex]{graphicx}

\title{Invariance before structure: two modes of emergent compression in representation learning}

\author{%
  Hardik Rajpal\\
  I-X Centre for AI in Science\\
  Department of Electrical and Electronics Engineering\\
  Imperial College London\\
  London, UK, SW7 2AZ \\
  \texttt{h.rajpal15@imperial.ac.uk} \\
  \And
  Dan Goodman \\
  Department of Electrical and Electronics Engineering\\
  Imperial College London\\
  London, UK, SW7 2AZ \\
  \texttt{d.goodman@imperial.ac.uk} \\
}

\begin{document}

\maketitle

\begin{abstract}
  Learned representations show striking invariance across biological and artificial neural networks, with different substrates and variations driven by drift, heterogeneity and initialisation. To understand this invariance, we propose Emergent Adaptive Representation Learning (EARL), a framework that distinguishes two modes of emergence in learned representations: state-space compression, where a learned representation gains predictive power over its own future beyond what the network's units can predict individually, and parameter-space compression, where networks with different micro-level parameters converge to the same representation. We formalise both as Fisher sensitivities and introduce ensemble-level information-theoretic measures for each. In a recurrent reservoir with a trained bottleneck, parameter-space compression rises within the first few epochs of training and partly independently of the task, whereas state-space compression rises about 30 epochs later and only when the network begins to learn the temporal structure of the task. These results suggest that invariance to the substrate and task-specific emergent structure are acquired in distinct phases of learning.
\end{abstract}

\section{Introduction}
From local synaptic plasticity to global gradient-based optimisation, biological and artificial neural networks learn through very different mechanisms \citep{lillicrap2020backpropagation,magee2020synaptic,rumelhart1986learning}, yet both these overparameterised systems are able to learn similar representations at the level of coordinated population activity \citep{saxena2019towards,vyas2020computation,zhang2016understanding}. In both, the same representation can be realised by very different micro-level configurations. In the brain, population representations remain stable despite representational drift, synaptic turnover and homeostatic recruitment among the underlying neurons \citep{driscoll2017dynamic,deitch2021representational,rule2019causes,noda2025homeostasis}, and similar representations arise across circuits with heterogeneous parameters, within and across individuals and species \citep{marder2006variability,safaie2023preserved}. In artificial networks, models trained from different initialisations and architectures often converge to similar representations when performing a common task \citep{li2015convergent,kornblith2019similarity}. Understanding how a learned representation becomes independent of its substrate is central to explaining why different brains and models represent the same stimuli similarly \citep{sucholutsky2023getting,huh2024platonic}.

We approach this question through \textit{emergence}, the property of complex systems whose collective behaviour cannot be reduced to that of their components \citep{anderson1972more,krakauer2026large}, which recent information-theoretic measures make quantifiable \citep{hoel2013quantifying,rosas2020reconciling,mediano2022greater}. Because emergence concerns information carried by the whole that is not present in any of its parts, it directly addresses the key property of learned representations implicated in substrate invariance. Such measures have characterised collective behaviour from bird flocks to ecosystems and organoids \citep{rosas2020reconciling,rajpal2025information,varley2025identification}, neural oscillations and conscious states \citep{rajpal2026quantifying,luppi2023reduced,milinkovic2026emergent}, and representations learned by deep networks \citep{kaplanis2023learning,mcsharry2024learning,yang2025finding}. In a classic reservoir computer, optimising the reservoir hyperparameters for prediction yields causally emergent representations \citep{tolle2026evolving}. In our previous work on a reservoir optimised for representation learning with a trained bottleneck, which we study here, the degree of emergence of the learned representation predicted generalisation to unseen rotations and dynamical systems \citep{rajpal2026emergent}. Distilling insights from various examples of emergence, we argue that emergent representations can arise through two distinct forms of compression, of the network's states and of its parameters. We introduce the Emergent Adaptive Representation Learning (EARL) framework, which formalises how both modes describe different aspects of generalisable and robust representation learning. We also propose empirically tractable information-theoretic measures for each. In a reservoir with a trained bottleneck, both measures show a transient, task-independent change during the initial fitting phase, followed by a robust rise later in training when the network learns the temporal structure of the task.

\section{Emergent Adaptive Representation Learning (EARL) framework}
\begin{wrapfigure}[26]{r}{0.27\linewidth}
  \centering
  \includegraphics[width=\linewidth]{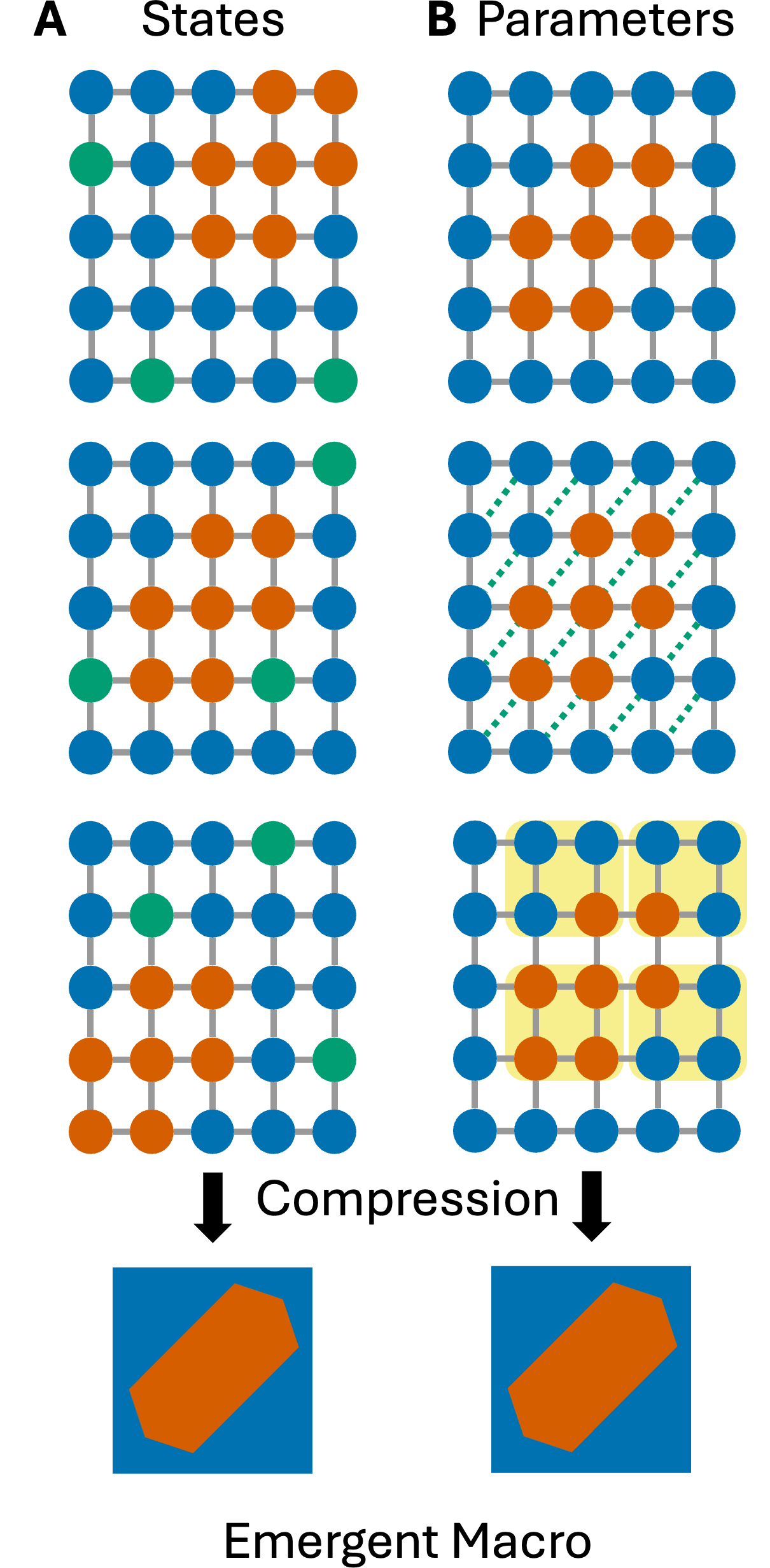}
  \caption{\protect\raggedright The two modes of compression. \textbf{A} State space: a stable macro configuration emerges as microscopic fluctuations are compressed away. \textbf{B} Parameter space: the same configuration emerges from different models of microscopic interactions.}
  \label{fig:modes}
\end{wrapfigure}
Emergent properties typically provide a compact description of the underlying system \citep{flack2017coarse,shalizi2001computational,krakauer2026large}. We distinguish two kinds of compression that can give rise to them (Fig.~\ref{fig:modes}).
\subsection{State-space compression}
In state-space compression, a coarse-grained description of the system's states predicts its future, or the task-relevant variables, better than its microscopic components do \citep{shalizi2001computational,rosas2020reconciling}. In physics, this corresponds to coarse-graining under the renormalisation group, which integrates out irrelevant degrees of freedom \citep{wilson1974renormalization,koch2018mutual}. In learning systems, it is related to the information bottleneck, which retains predictive information while discarding the rest of the input \citep{tishby2000information,achille2018emergence} and has been linked to better in-task generalisation and robustness to input noise \citep{alemi2016deep,kawaguchi2023does}.

\subsection{Parameter-space compression}
In parameter-space compression, macroscopic behaviour depends on only a few combinations of the microscopic parameters: many models in physics and biology are sensitive to a few stiff parameter combinations and insensitive to the rest \citep{gutenkunst2007universally}. Such overparameterised systems often exhibit universality, where many microscopic models share one macroscopic description \citep{machta2013parameter}. In the brain, this corresponds to degeneracy, where disparate circuit parameters and continual synaptic turnover yield the same function \citep{edelman2001degeneracy,marder2006variability,attardo2015impermanence}. In artificial networks, it underlies the convergence of representations across initialisations and model architectures \citep{li2015convergent,kornblith2019similarity}.
\subsection{Formalism}
Consider a neural network whose micro-state $X_t \in \mathbb{R}^N$, the activity of its $N$ units, evolves under micro-level parameters $\theta$, and a learned representation $V_t = f_w(X_t)$ with trainable weights $w$ that encodes a task-relevant latent variable $L_t$, such as the variables generating a simulated dataset or relevant behavioural variables recorded in an experiment. Here $\theta$ collects the parameters that define the substrate and are held fixed across realisations of the network, such as architectural hyperparameters or fixed weights. Both modes can be phrased as Fisher sensitivities of the representation (Appendix~\ref{app:fisher}); we instead use mutual-information analogues, which can be estimated across an ensemble from recorded activity without access to gradients \citep{brunel1998mutual}.

\section{Measuring state-space and parameter-space compression}\label{sec:measures}

\subsection{State-space compression as causal emergence}
A representation that compresses the state space should predict its own future better than its micro-variables do individually. We therefore quantify state-space compression with the practical criterion for causal emergence of Rosas et al. \citep{rosas2020reconciling}, applied to the learned representation:
\begin{equation}
    \Psi = I(V_t; V_{t+1}) - \sum_{i=1}^{N} I(X_t^i; V_{t+1})
\end{equation}
The first term is the self-predictability of the representation and the second sums the information each unit carries about the representation's future. Following \citet{sas2025improved}, we add a second-order correction, which adds back the redundancy shared by pairs of units and so minimally accounts for the information double-counted in the micro sum; this correction has previously been applied to neural recordings \citep{rajpal2026quantifying}. With many units, higher-order redundancies are not accounted for. All terms are Gaussian mutual-information estimates (in nats), computed per sequence and averaged over sequences and models. $\Psi > 0$ is sufficient but not necessary for emergence \citep{rosas2020reconciling,mediano2022greater}. Because the micro sum still over-counts information shared by many units, $\Psi$ is typically negative in large networks, and we interpret relative changes: an increase in $\Psi$ means the representation's future is more predictable from the representation itself and less attributable to any single unit.

\subsection{Parameter-space compression as degeneracy}
Parameter-space compression asks the converse question: whether networks with different micro-level parameters $\theta$ arrive at the same representation. Following the idea of parameter-space compression of Machta et al. \citep{machta2013parameter}, we train an ensemble across $\theta$ and measure how much information a task-relevant feature of the representation, $F(V)$, retains about $\theta$, relative to a matched feature of the micro-level activity, $F(X)$:
\begin{equation}
    \Lambda = 1 - \frac{I\big(\theta; F(V) \mid \nu\big)}{I\big(\theta; F(X) \mid \nu\big)}
\end{equation}
We take $F(V) = I(V;L)$, the Gaussian mutual information between the learned and the true latent, which is invariant to invertible linear transformations of $V$ and $L$ and defined for any task with a latent variable. The micro feature $F(X)$ consists of summary statistics of fixed random projections of the micro-state; if the micro-level dynamics are not trained, the denominator is constant across training. Here $\nu$ collects nuisance variables shared across $\theta$, namely the input sequence and the initialisation of the trained weights; both features are centred across $\theta$ within each value of $\nu$, so $\Lambda$ measures information about $\theta$ beyond what $\nu$ already determines. $\Lambda = 1$ means the representation carries no detectable information about $\theta$; $\Lambda = 0$ means it carries as much as the micro-level activity does.

We estimate each $I(\theta;\cdot)$ with a held-out Barber--Agakov lower bound \citep{barber2003im}, treating the sampled settings of $\theta$ as equiprobable classes and fitting and scoring on disjoint sequences. Every estimate is reported alongside a null computed with shuffled $\theta$ labels. The number of random projections and the choice of estimator shift the level of $\Lambda$ but not its trend (Appendix~\ref{app:estimator}).

\section{Model setup}\label{sec:model}
We use the reservoir--bottleneck architecture from our previous work \citep{rajpal2026emergent}: a fixed random recurrent reservoir of $N=500$ leaky-tanh units, a trainable 10-dimensional linear bottleneck $Z_t = W_z X_t + b_z$, and a ridge-regression readout that predicts the next input using the bottleneck states (Appendix~\ref{app:model}). In this model, $X_t$ are the $N=500$ reservoir states, $V_t$ the top three principal components of the bottleneck activity $Z_t$ (fitted on the training sequences), $L_t$ the true three-dimensional attractor state, and $\theta$ the reservoir hyperparameters. The micro feature $F(X)$ is the time mean and log-variance of 24 fixed random projections of $X_t$; because the reservoir is fixed, $I(\theta;F(X))$ is constant over training. Inputs are random orthogonal 10-dimensional embeddings of three-dimensional attractors; the bottleneck is trained by backpropagation through the ridge readout to minimise one-step prediction error on 12 training sequences (Sprott B, R\"ossler, Lorenz), and evaluated on 12 sequences with unseen rotations of the same attractors and 12 sequences of held-out attractors (Chen, Sprott A). Each ensemble shares one reservoir draw and varies $\theta$ (spectral radius, input scaling, bias scaling, leak rate and ridge penalty) over 100 settings drawn log-uniformly within $\pm30\%$ of the Optuna optimum \citep{akiba2019optuna}, with sparsity fixed. Every setting is trained from the same four bottleneck initialisations. We report 20 replicate ensembles (5 reservoir draws $\times$ 4 data seeds), trained for 100 epochs on noise-free input (noisy input shown in Appendix~\ref{app:noise}), and two matched controls with 10 replicates each, in which the training target is either time-permuted within its sequence or taken from another sequence.

\section{Results}
\begin{figure}[t]
\centering
\includegraphics[width=\linewidth]{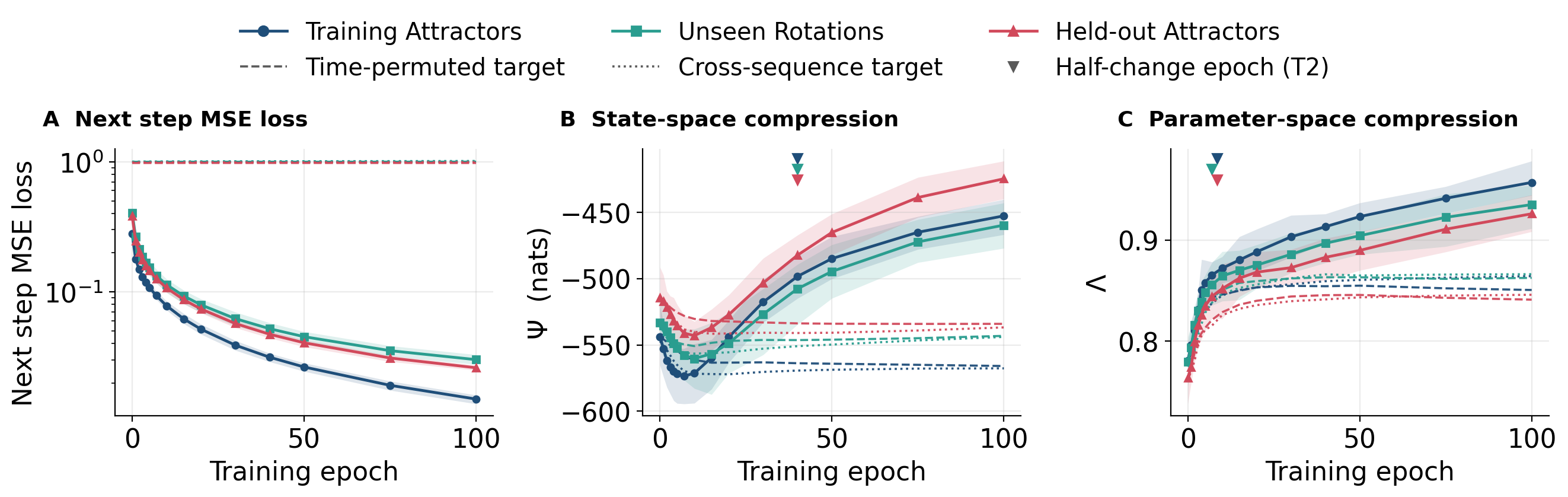}
\caption{Learning dynamics on noise-free input. \textbf{A} One-step prediction error. \textbf{B} State-space compression $\Psi$. \textbf{C} Parameter-space compression $\Lambda$. Solid: mean of 20 replicate ensembles (band: 10th--90th percentile) on training attractors, unseen rotations of the training attractors and held-out attractors; dashed/dotted: time-permuted and cross-sequence target controls (10 matched replicates each). Triangles: half-change epochs (from epoch 0 for $\Lambda$, from the minimum for $\Psi$).}
\label{fig:main}
\end{figure}

Training reduces the one-step prediction error monotonically on all three evaluation sets, while the learned representation becomes increasingly aligned with the true latent (Fig.~\ref{fig:main}A; Appendix~\ref{app:alignment}). State-space compression is non-monotonic (Fig.~\ref{fig:main}B): $\Psi$ first decreases, reaching its minimum by about epoch 10, and then rises for the rest of training, reaching half of its rise by about epoch 40 on all three sets, consistent with our previous findings \citep{rajpal2026emergent}. As expected for $N=500$ units, $\Psi$ is negative throughout.

Parameter-space compression instead rises from the start of training (Fig.~\ref{fig:main}C): $\Lambda$ increases from about 0.78 to 0.96 on the training sequences ($\Delta\Lambda = 0.177 \pm 0.020$, mean $\pm$ s.d.; similar on the other sets), and increases on every set (two-sided sign test across replicates, $p = 1.9\times10^{-6}$). $\Lambda$ reaches half of its total change by about epoch 8 and $\Psi$ half of its rise by about epoch 40 (triangles in Fig.~\ref{fig:main}B,C), so parameter-space compression leads state-space compression by $\Delta T \approx 32$ epochs (difference of median half-change epochs), and $\Lambda$ is never later than $\Psi$ (sign test, $p \le 1.5\times10^{-5}$).

Both controls reproduce the early phase but not the increase in later stages. With a time-permuted or cross-sequence target, $\Psi$ still dips and $\Lambda$ still rises to about 0.85, but $\Psi$ does not recover and $\Lambda$ plateaus (Fig.~\ref{fig:main}B,C, dashed and dotted lines). Over training, models trained on the true target gain more than their matched controls (paired sign test, $p = 0.002$ for each control and set): by about 115 nats in $\Psi$ and 0.09--0.10 in $\Lambda$ on the training sequences. Results with noisy input are qualitatively the same (Appendix~\ref{app:noise}). Together, these results separate the two modes: both show a task-independent early phase, in which $\Psi$ dips and $\Lambda$ makes its first rise with or without the task, whereas the recovery of state-space compression and the continued gain in parameter-space compression appear only when the network learns the temporal structure of the task.

\section{Discussion}
We show that training may first make the representation invariant to the substrate, on which task learning then builds a self-predictive, distributed representation. Because $\Lambda$ compares models through what their representations encode rather than by aligning the representations directly, it also offers an information-theoretic measure of representational universality across substrates.

Both measures are relative: $\Psi$ is negative throughout and its trend depends on the temporal sampling of the data, and $\Lambda$ requires a task-relevant latent and measures invariance across models rather than the robustness of a single model. The measures extend to other tasks by replacing $V_{t+1}$ in $\Psi$ and $F(V)$ in $\Lambda$ with task-appropriate quantities. We have previously observed the non-monotonic trajectory of $\Psi$ in hippocampal and prefrontal recordings during learning \citep{rajpal2026emergent}; measuring parameter-space compression across animals, and in datasets with representational drift \citep{driscoll2017dynamic,deitch2021representational}, is a natural next step.

\begin{ack}
H.R. was funded by the Eric and Wendy Schmidt AI in Science Postdoctoral Fellowship, supported by Schmidt Sciences, LLC.
\end{ack}

{\small
\bibliographystyle{unsrtnat}
\bibliography{unireps_2026}
}

\clearpage
\appendix
\section{Model architecture}\label{app:model}
\begin{figure}[!h]
\centering
\includegraphics[width=0.8\linewidth]{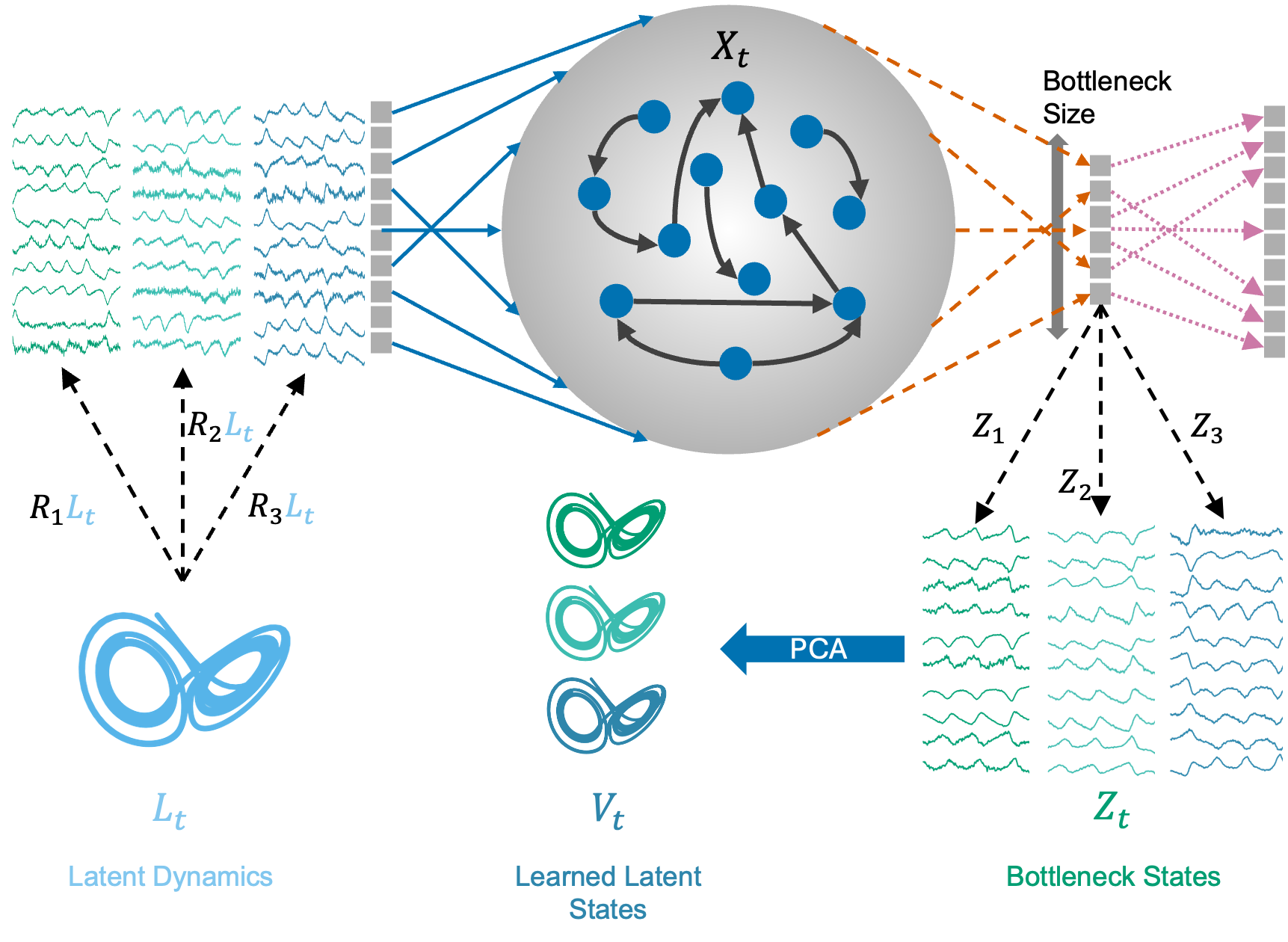}
\caption{Reservoir--bottleneck architecture, adapted from our previous work \citep{rajpal2026emergent}. Random orthogonal embeddings of the attractor latent $L_t$ drive a fixed random reservoir with states $X_t$; a trainable linear bottleneck maps $X_t$ to $Z_t$, whose top three principal components form the learned representation $V_t$; a ridge readout predicts the next input. The bottleneck is the only trained component.}
\end{figure}

\section{Fisher-information formulation}\label{app:fisher}
We treat the micro-state as evolving under a transition kernel $p_\theta(X_{t+1}\mid X_t)$, with external input treated as a source of stochasticity and marginalised, and read out the representation with additive noise, $V_t = f_w(X_t) + \varepsilon_t$, $\varepsilon_t \sim \mathcal{N}(0,\sigma^2 I)$; the noise makes the conditional densities below well defined.

\paragraph{State space.} The sensitivity of the representation's future to the current micro-state is
\begin{equation}
    F_\text{state} = \mathbb{E}\big[\nabla_{X_t}\log p_\theta(V_{t+1}\mid X_t)\,\nabla_{X_t}\log p_\theta(V_{t+1}\mid X_t)^\top\big] = \sigma^{-2}\,\mathbb{E}\big[J_t^\top J_t\big], \qquad J_t = \frac{\partial V_{t+1}}{\partial X_t}.
\end{equation}
This mirrors the micro term of $\Psi$. A constant representation trivially has $F_\text{state}=0$, so compression means that sensitivity concentrates on the few directions of $X_t$ that predict the target while $I(V_t;V_{t+1})$ is preserved.

\paragraph{Parameter space.} The sensitivity of the encoding of the latent to the micro-level parameters is
\begin{equation}
    F_\text{param} = \mathbb{E}\big[\nabla_\theta \log p_{\theta,w^*(\theta)}(V_t\mid L_t)\,\nabla_\theta \log p_{\theta,w^*(\theta)}(V_t\mid L_t)^\top\big], \qquad \nabla_\theta = \partial_\theta + \Big(\frac{\partial w^*}{\partial\theta}\Big)^{\!\top}\partial_w.
\end{equation}
We condition on the true latent $L_t$, matching $F(V)=I(V;L)$, rather than on $X_t$, which would block the dependence of $V$ on $\theta$ through the micro-state dynamics. The total derivative includes the adaptation of the trained weights $w^*(\theta)$: degeneracy means that training compensates for changes in $\theta$, making this Fisher small relative to that of the micro-level activity. $\Lambda$ is an ensemble-level information-theoretic analogue of this comparison. In the low-noise limit Fisher and mutual information are related \citep{brunel1998mutual}, which motivates the substitution in the main text. Unlike sloppy-model analyses \citep{gutenkunst2007universally,machta2013parameter}, we do not examine the eigenvalue spectrum of $F_\text{param}$.

For the reservoir model of Section~\ref{sec:model}, $X_t$ are the reservoir states, $w$ the bottleneck weights, $\theta$ the reservoir hyperparameters, and $L_t$ the true attractor state.

\section{Estimator details and robustness checks}\label{app:estimator}
\begin{figure}[!h]
\centering
\includegraphics[width=\linewidth]{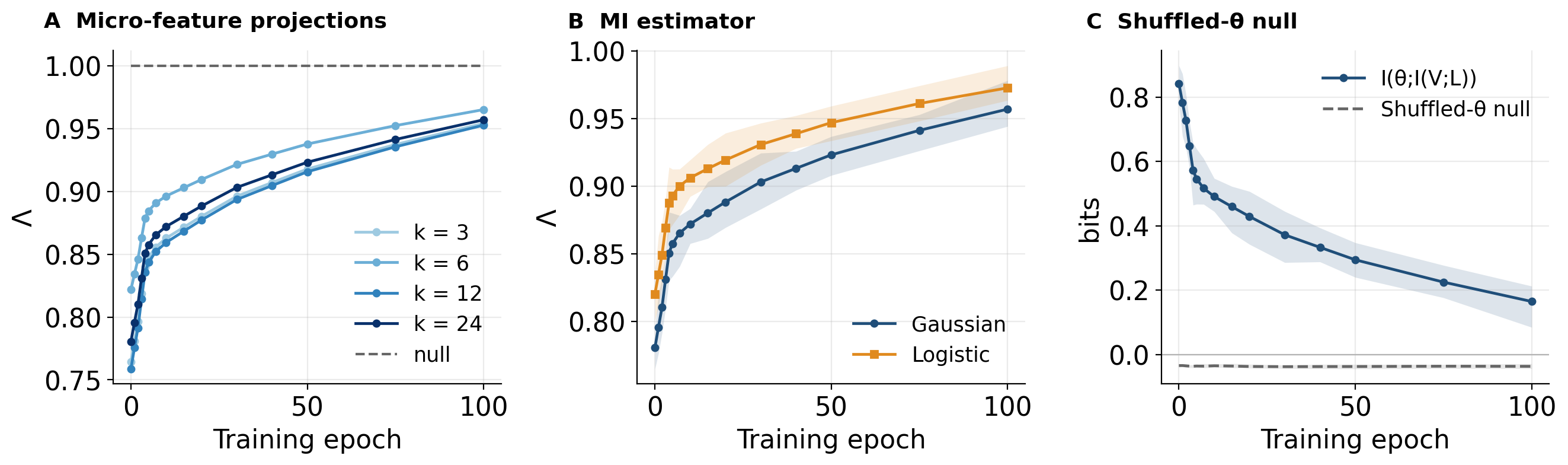}
\caption{Robustness of parameter-space compression (noise-free input, training sequences). \textbf{A} $\Lambda$ with the micro feature $F(X)$ built from $k$ random projections of the reservoir states ($k=24$ in the main text); dashed line: shuffled-label null, $\Lambda = 1$. \textbf{B} $\Lambda$ with the Gaussian Barber--Agakov bound used in the main text and with a multinomial-logistic bound. \textbf{C} The numerator of $\Lambda$, $I(\theta;F(V))$ with $F(V)=I(V;L)$, and the same bound with shuffled $\theta$ labels. Both choices shift the level of $\Lambda$ but not its rise, and $I(\theta;F(V))$ stays above its null, so the approach of $\Lambda$ towards 1 is not an estimator floor.}
\label{fig:robust}
\end{figure}

The class-conditional model in the Barber--Agakov bound is Gaussian with covariance shrinkage 0.5; we fit both a plain and a rank-normal Gaussian and report the larger bound, since both are valid lower bounds. $\Lambda$ is reported only when $I(\theta;F(X)) > 0.2$ bits, a negative held-out $I(\theta;F(V))$ is clipped to zero, and $\Lambda = 1 - \max(I(\theta;F(V)),0)/\max(I(\theta;F(X)), I(\theta;F(V)))$, so that $\Lambda \in [0,1]$. The choice of micro feature and of bound changes the level of $\Lambda$ but not its rise over training, and the information about $\theta$ retained by the representation stays above the shuffled-label null throughout (Fig.~\ref{fig:robust}).

\clearpage
\section{Alignment with the true latent}\label{app:alignment}
\begin{figure}[!h]
\centering
\includegraphics[width=0.75\linewidth]{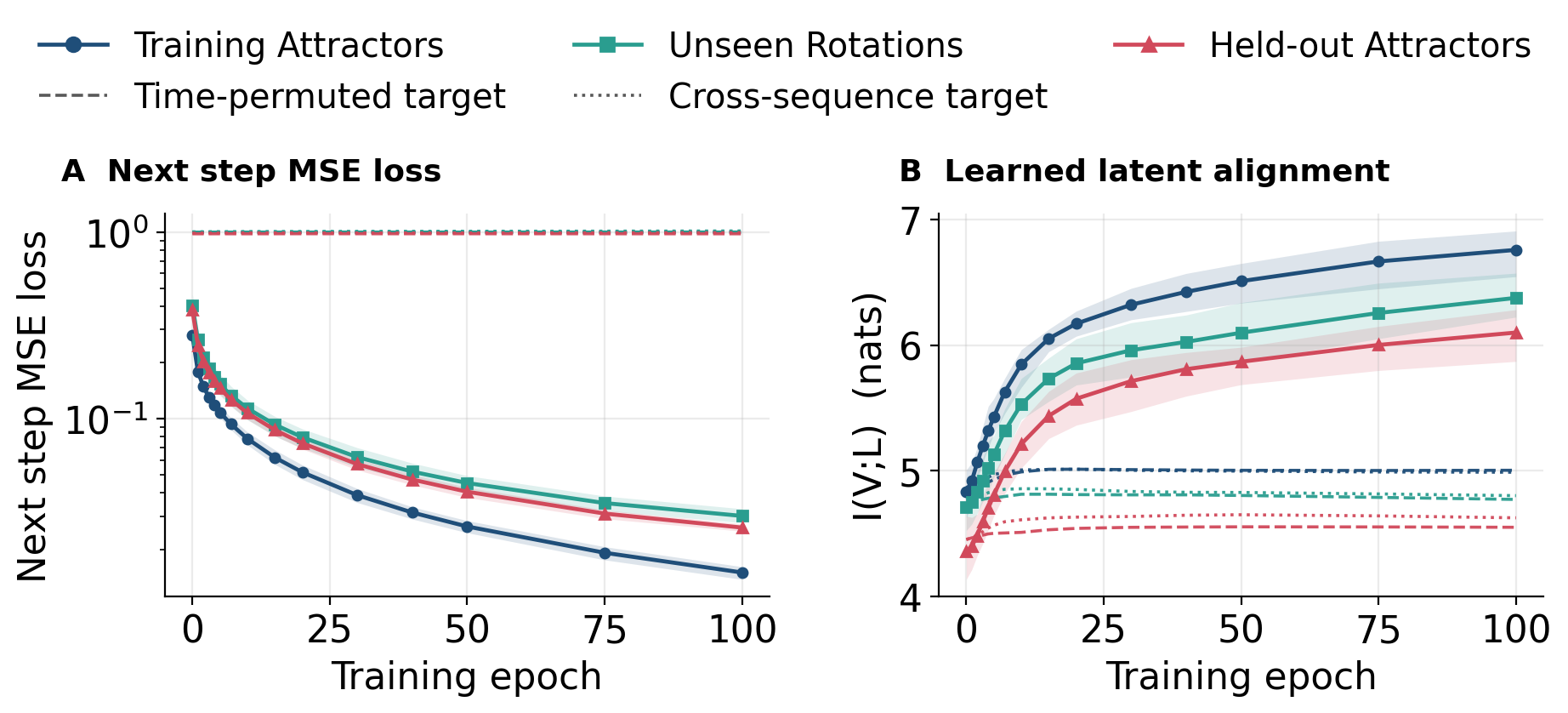}
\caption{Performance and alignment for noise-free input, with lines and bands as in Fig.~\ref{fig:main}. \textbf{A} One-step prediction error (as Fig.~\ref{fig:main}A). \textbf{B} Alignment of the learned with the true latent, $I(V;L)$.}
\label{fig:alignment}
\end{figure}

\section{Noisy input}\label{app:noise}
\begin{figure}[!h]
\centering
\includegraphics[width=\linewidth]{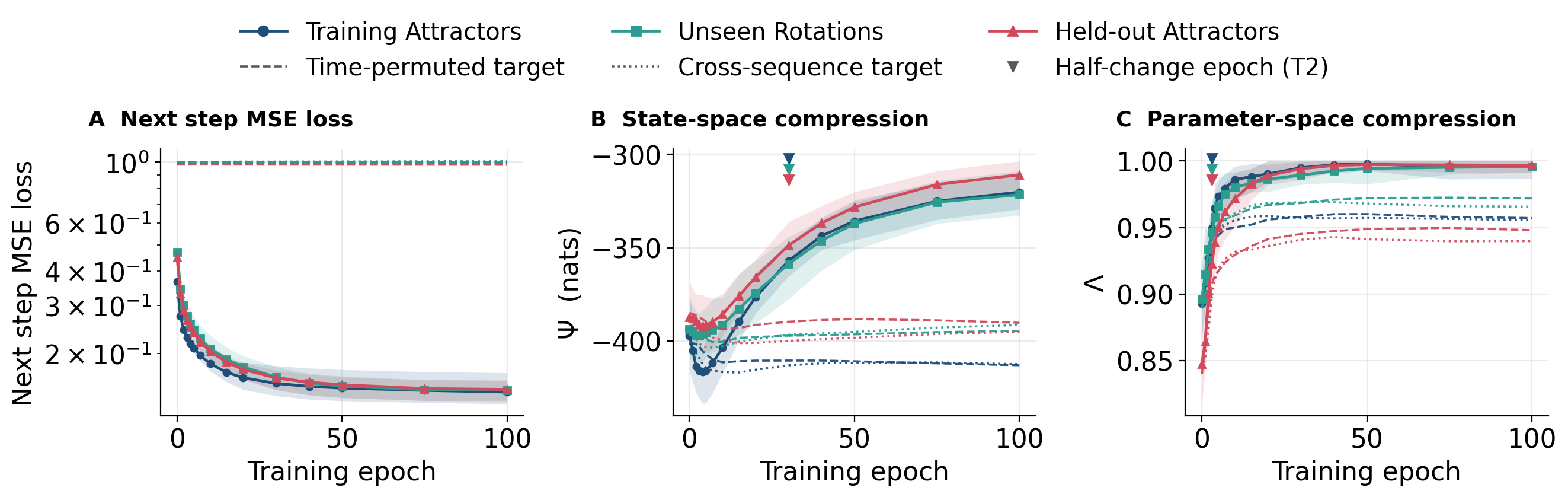}
\caption{As Fig.~\ref{fig:main}, for models trained and evaluated on noisy input ($\sigma_y = 0.1$). $\Lambda$ approaches 1 in trained runs and plateaus near 0.95 in controls; the ordering of the two modes and the differences from the controls are preserved.}
\end{figure}

\end{document}